\documentclass[letterpaper,10pt,conference]{ieeeconf}

\usepackage[acronym,nonumberlist]{glossaries}
\usepackage{amsmath}
\usepackage{amssymb}

\usepackage[ruled,vlined]{algorithm2e}

\usepackage{hyperref}
\newcommand\rurl[1]{%
\texttt{\href{http://#1}{\nolinkurl{#1}}}%
}

\newacronym{ood}{OoD}{Out-of-Distribution}
\newacronym{cnn}{CNN}{convolutional neural network}
\newacronym{aspp}{ASPP}{Atrous Spatial Pyramid Pooling}
\newacronym{nn}{NN}{neural network}
\newacronym{dnn}{DNN}{deep neural network}
\newacronym{mlp}{MLP}{multilayer perceptron}
\newacronym{iou}{IoU}{intersection over union}
\newacronym{miou}{mIoU}{mean intersection over union}
\newacronym{bce}{BCE}{binary cross-entropy}
\newacronym{sota}{SOTA}{state-of-the-art}
\newacronym{sotif}{SOTIF}{``Safety of the Intended Functionality''}
\newacronym{mim}{MIM}{Masked Image Modeling}
\newacronym{mlm}{MLM}{Masked Language Modeling}
\newacronym{candr}{C\&R}{crop-and-resize}
\newacronym{dum}{DUM}{deterministic uncertainty method}
\newacronym{vit}{ViT}{vision transformer}

\usepackage{pifont}

\newcommand{\ftheta}{$\mathtt{f}_{\theta}$\xspace}
\newcommand{\fphi}{$\mathtt{f}_{\phi}$\xspace}

\usepackage[binary-units]{siunitx}
\usepackage[labelformat=simple]{subcaption}
\DeclareCaptionLabelSeparator{periodspace}{.\quad}
\usepackage{textcomp}
\usepackage{eucal}
\usepackage{bbm}
\usepackage{graphicx}
\usepackage{multirow}

\usepackage{cleveref}
\crefname{table}{Tab.}{Tabs.}
\crefname{figure}{Fig.}{Figs.}
\crefname{section}{Sec.}{Secs.}
\crefname{equation}{Eq.}{Eqs.}

\usepackage{bbding}

\IEEEoverridecommandlockouts

\usepackage{cite}

\begin{document}

\title{\Large \bf Masked $\gamma$-SSL: Learning Uncertainty Estimation via Masked Image Modeling}
\author{David S. W. Williams, Matthew Gadd, Paul Newman and Daniele De Martini\\
Oxford Robotics Institute, Dept. of Engineering Science, University of Oxford, UK.\\\texttt{\{dw,mattgadd,pnewman,daniele\}@robots.ox.ac.uk}
\thanks{This work was supported by EPSRC Programme Grant ``From Sensing to Collaboration'' (EP/V000748/1).
}
}
\maketitle


\begin{abstract}
This work proposes a semantic segmentation network that produces high-quality uncertainty estimates in a single forward pass.
We exploit general representations from foundation models and unlabelled datasets through a \gls{mim} approach, which is robust to augmentation hyper-parameters and simpler than previous techniques.
For neural networks used in safety-critical applications, bias in the training data can lead to errors; therefore it is crucial to understand a network's limitations at run time and act accordingly.
To this end, we test our proposed method on a number of test domains including the SAX Segmentation benchmark, which includes labelled test data from dense urban, rural and off-road driving domains.
The proposed method consistently outperforms uncertainty estimation and \gls{ood} techniques on this difficult benchmark.
\end{abstract}

\glsresetall

\section{Introduction} \label{sec:introduction}

\begin{figure}[]
\centering
\includegraphics[width=0.7\columnwidth]{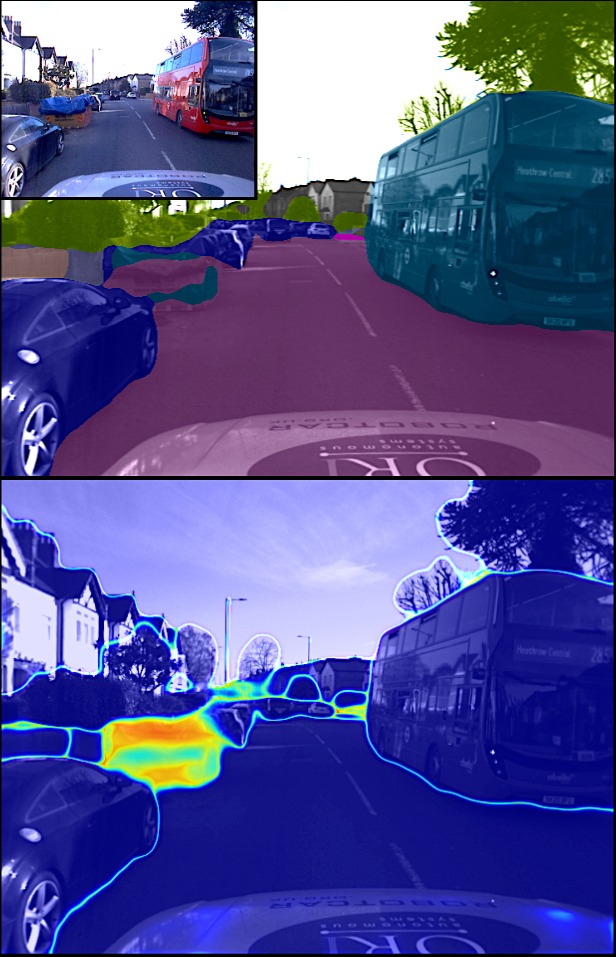}
\caption{
Our proposed method jointly performs high-quality semantic segmentation (top) \textit{and} pixel-wise uncertainty estimation (bottom) on an image (top, top left) from the SAX London test dataset.
Note a dumpster (an undefined semantic class) is \textit{inaccurately} segmented, however our network is correspondingly uncertain (bottom, yellow and red), while the rest of the segmentation is accurate and certain (bottom, blue).
See \cref{appendix} for more qualitative results.
}
\label{fig:system_overview}
\vspace{-15pt}
\end{figure}

Semantic segmentation gives robots a useful semantic understanding of their surroundings. 
Segmentation networks are typically trained with data annotated with the relevant semantic concepts; however, when presented with instances of classes with a different appearance or instances of a different class, they are prone to making mistakes.
In the safety-critical context of perception in robotics, this is often unacceptable, and thus, methods must be developed to help neural networks understand their limitations.
This work presents a method that produces high-quality uncertainty estimates for a semantic segmentation task, focusing on the benefits of leveraging general representations from foundation models.
In addition, we exploit easy-to-collect, unlabelled domain-specific datasets.
Specifically, given a labelled training dataset in one domain (a.k.a. the \emph{source domain}), we are interested in mitigating the segmentation error rate on a distributionally-shifted unlabelled domain (a.k.a. the \emph{target domain}).

Recently, it has become effective to perform self-supervised training on diverse image datasets which try to approximate the set of all natural images.
The intent is to learn a general semantic representation of the provided unlabelled images.
A primary benefit is that these models can be fine-tuned to solve specific image-based perception tasks, giving a particularly large boost in performance to tasks with small amounts of data.
If we view distributional shifts as a problem of insufficient labelled data, this suggests that they can be ameliorated by leveraging these network representations.

In this work, we define our task of interest as the semantic segmentation of images from a source domain for which labelled training images are provided.
The encoder of a segmentation network is initialised with a general representation -- namely DINOv2 \cite{dinov2} -- and then the entire network is specialised to solve the task at hand.
A key question is then: as a result of this necessary model specialisation, how much does the quality of uncertainty estimation degrade on images from different target domains?

This work investigates this question and presents a method that uses \gls{mim} on unlabelled images to investigate a given segmentation network's representation of distributionally-shifted data \textit{during training}, in order to learn high-quality uncertainty estimation.

\Gls{mim} has recently been used to learn representations from images in a self-supervised manner by removing patches of the input image and having the network predict the information content of these masked regions.
This task requires a detailed understanding of the unmasked patches, and the semantic interactions between masked and unmasked patches, and is thus also predictive of network robustness.

When presented with a masked, unfamiliar image, if the network cannot extract sufficient salient information from the unmasked patches to accurately segment the masked patches, the segmentation is likely to differ from that of the same image, but unmasked.
Without the requirement of labelled images, this gives us the signal required to discover the limits of the network, and how image appearance relates to segmentation quality.


This work is based on our previous work \cite{gammassl}, beyond which this work's contributions are as follows:
\begin{itemize}
    \item A \gls{mim}-based method for learning uncertainty estimation from unlabelled data, which is simpler, produces higher quality uncertainty estimates and with less sensitivity to augmentation hyperparameters;
    \item An empirical investigation of the effects of distributional shift on networks with general image representations.
\end{itemize}


\section{Related Work} \label{sec:related_work}
\subsection{Epistemic Uncertainty Estimation}
As epistemic uncertainty is defined as inherent to the model, and not the data, this set of literature concerns itself with modeling the distribution of model parameters.
A full Bayesian treatment is intractable, however a number of possible approximations are made in \cite{bayes_by_backprop, practical_vi, louizos2017multiplicative}, and notably in Monte Carlo Dropout networks \cite{dropout}.
Uncertainty estimation is performed by sampling sets of parameters, segmenting with each, and then evaluating the segmentation consistency.
An alternative that produces high-quality uncertainty estimates is to model this model distribution with an ensemble \cite{ensembles}.



The major drawback to these methods for robotics is that they require multiple forward passes of a network at run-time, greatly increasing the latency, which is often unacceptable in safety-critical contexts.
Our proposed method is therefore designed to use only a single forward pass at run-time.

\subsection{Aleatoric Uncertainty Estimation}
For aleatoric uncertainty estimation, the uncertainty is inherent to the data, and not the model.
The focus is therefore to learn the appearance of regions that are likely to result in error. 
\cite{what_uncertainties,probably_unknown} parameterise the network output with a Gaussian distribution, such that the network is softly-constrained to either estimate the quantity of interest, or to output a large variance.
\cite{light_prob_nets} parameterises the features from each layer as a distribution, using assumed density filtering.

The above uses labelled images, such that ground-truth error is known during training.
However \cite{aleatoric_geo_stable} uses unlabelled images to solve both a geometric matching problem and uncertainty estimation by parameterising the output as an exponential distribution.
This work, along with \cite{gammassl}, also leverages unlabelled images to learn uncertainty estimation, but for a semantic segmentation task.



\subsection{Single-Pass Uncertainty Methods}
\Glspl{dum}~\cite{dum_baseline,duq,sngp} estimate uncertainty in a single forward pass with the use of spectral normalisation layers~\cite{spectral_norm}, which restrict the Lipschitz constant of the model with the hope that large semantic differences in the input lead to large distances in feature space.
Alternatively, \cite{variance_prop} injects noise during model training, and calculates the approximate feature covariances as a measure of epistemic uncertainty.
\cite{orthonormal_certificates} trains a set of orthonormal linear layers on top of a feature extractor, such that each feature from the training dataset is mapped to zero - thus epistemic uncertainty is represented by the average output of linear layers.
\cite{deepvib_uncertainty} shows that regularisation from the Deep Variational Bottleneck method~\cite{deepvib} allows for improved single-pass uncertainty estimation.

Much like the cited \glspl{dum} and \cite{deepvib_uncertainty}, this work turns uncertainty estimation into a representation learning problem, which is solved by refining a general feature representation using \gls{mim} uncertainty training and unlabelled \gls{ood} data.




\subsection{Out-of-Distribution Detection}

\gls{ood} detection methods find images that are distinct from the distribution defined by a given dataset.
A common procedure is to train a network on this dataset, then freeze it and develop an inference method that leverages the learned representation.
\cite{maxsoftmax} calculates the max softmax score, \cite{mahalanobis} uses a Gaussian Mixture model and \cite{vim} calculates a score from both features and logits.

Another form of \gls{ood} detection explicitly trains a network to separate labelled training and \gls{ood} data, by training on a dataset of curated \gls{ood} images, \cite{prior_nets, outlierexposure}.
However, \cite{fool_me_once} and \cite{ood_gan}, show that the greater the difference between the source and \gls{ood} datasets, the worse the \gls{ood} detection.
We take inspiration from this, and use a dataset which is composed of images with in-distribution, near-distribution and out-of-distribution instances all within the same images.

\subsection{Masked Image Modelling} \label{subsec:mim}
Recently, \gls{mim} has been utilised in order to learn a representation from diverse image data in a self-supervised manner, in conjunction with the Transformer architecture~\cite{transformer}.
\cite{mae} trains a \gls{vit} to reconstruct the RGB values for masked patches, while \cite{dinov2, ibot} maximise the consistency between extracted masked and unmasked features.
\cite{mae} masks randomly, while \cite{sem_mae} masks semantic parts, and \cite{adversarial_masking} learns a masking policy.
These methods are ultimately judged on the network's accuracy on diverse tasks, for which the network is fine-tuned.
Instead of model pre-training, this work uses \gls{mim} to fine-tune a model trained with the above methods, focusing purely on producing high-quality uncertainty estimates.



\section{Preliminaries and Notation} \label{sec:system_design}

\begin{figure*}[ht]
\centering
\includegraphics[width=0.7\textwidth]{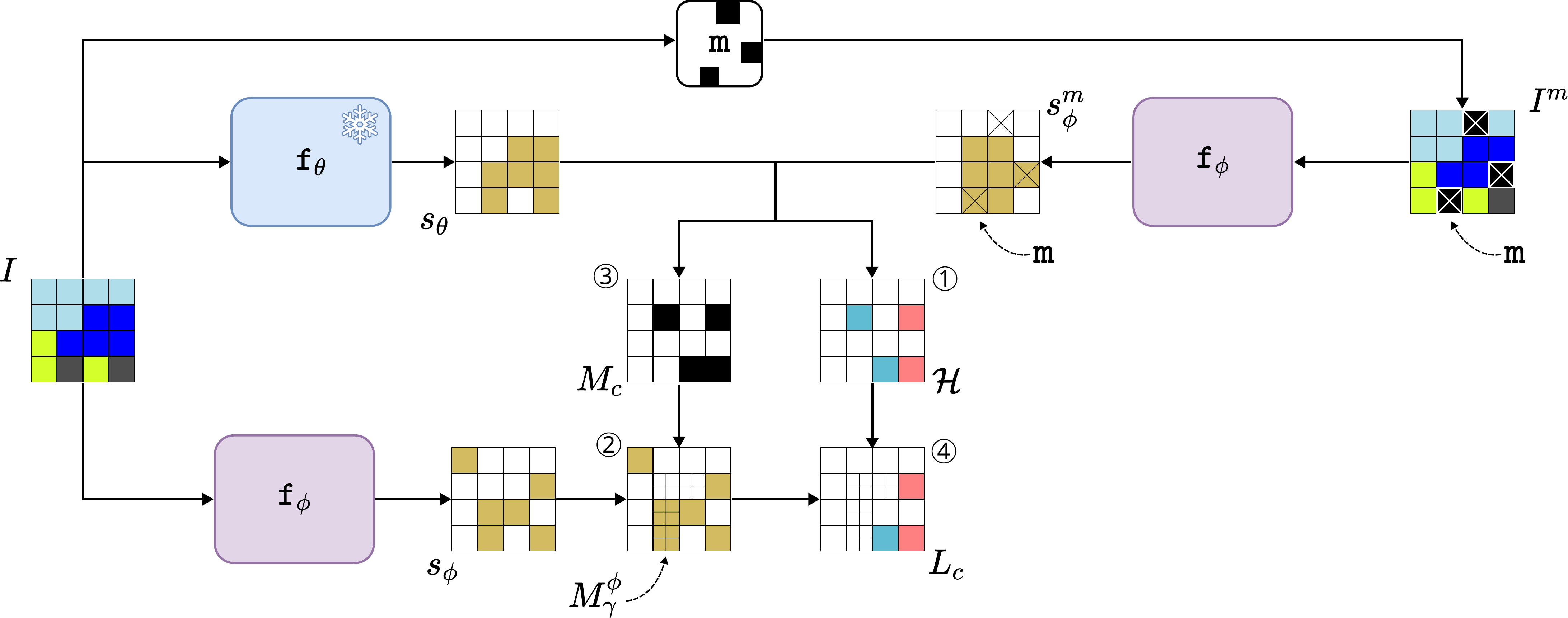}

\caption{
\label{fig:network_diagram}
\textit{Overview of our uncertainty training framework.}
It involves frozen network \ftheta (blue) and the network being trained \fphi (purple).
\ftheta has been trained to perform semantic segmentation with images from a labelled domain.
Using unlabelled images in a different domain, regions of likely segmentation error are found by comparing masked segmentation $s_{\phi}^{m}$ (masking denoted by $\boxtimes$) and $s_\theta$.
Loss $L_c$ refines the uncertainty estimates of \fphi by minimising the soft consistency $\mathcal{H}$ \ding{172}, but only for regions where \fphi is certain, implemented via an uncertainty masking procedure by binary mask $M_\gamma^{\phi}$ \ding{173} (denoted by $\boxplus$).
The threshold for $M_\gamma^{\phi}$ is calculated using a hard consistency mask \ding{174}, giving the final loss \ding{175}.
}
\vspace{-15pt}
\end{figure*}

%

\subsection{Semantic Segmentation and Uncertainty Estimation}
Semantic segmentation requires a model, $\mathtt{f}$, to assign a class to each pixel of an RGB image, $I \in \mathbb{R}^{3 \times H \times W}$, resulting in a segmentation mask, $s \in \mathbb{R}^{K \times H \times W}$, where $K = | \mathcal{K} |$, and $\mathcal{K}$ is the set of defined semantic classes.
The model, $\mathtt{f} = \mathtt{D} \circ \mathtt{E}$, is composed of a feature encoder, $\mathtt{E}$, and a segmentation decoder, $\mathtt{D}$.

Uncertainty estimation for this task requires a model to also produce a pixel-wise score, representing the likelihood of a pixel being assigned to the wrong class.
Therefore, $\mathtt{f}$ also produces a confidence score mask $u \in \mathbb{R}^{H \times W}$.

One method for calculating this score from a segmentation network is to consider the categorical distribution for each pixel (by applying the $\mathrm{softmax}$ function to each pixel location as the final layer) and to calculate the maximum over the probabilities in order to assign a class to each pixel.
This value gives us the network confidence, calculated as $u = \mathrm{max} \circ \mathrm{softmax} \circ \mathtt{f}(I) $ and, for completeness, $s = \operatorname{argmax}{~\mathtt{f}(I)}$.
Given a threshold $\gamma$, we can then calculate a binary confidence mask $M_\gamma \in \{0,1\}^{H \times W}$ which is defined such that for the pixel $i$:
\begin{equation}\label{eq:mgamma}
    M_\gamma^{(i)} = u^{(i)} > \gamma
\end{equation}
Pixels which are \texttt{1} in this mask have confident class predictions, while \texttt{0} have uncertain class predictions.

Unless otherwise stated, uncertainty is estimated in this way, and this is the quantity that we seek to improve with uncertainty training in this work.

\subsection{Overview of the Proposed Approach}

The method in this work is part of a three-step training process: pre-training, task learning, and uncertainty training; these are described in more detail in the following.
This work proposes a method for the final of the three steps.

\subsubsection{Pretraining}
The first step trains the encoder to produce a general feature representation of natural images.
This relates to the pre-training of the encoder $\mathtt{E}$ component of $\mathtt{f}_{\theta}$ and $\mathtt{f}_{\phi}$ to build a good initialisation for the further specialisation of $\mathtt{f}_{\theta}$ and $\mathtt{f}_{\phi}$ during the next steps.

The benefits of this are as follows: (1) this step is only done once, allowing any subsequent task learning to be achieved more quickly and easily (2) if only a relatively small labelled dataset is available, the segmentation quality is likely to be better in the source domain and (3) also in the target domains; (4) the uncertainty estimation will be better in both the source and target domains (see \cref{subsec:results_weight}).

\subsubsection{Task Learning}
The second step specialises the model to perform the task of interest. 
In this work, labelled source domain images are used to maximise semantic segmentation quality in a supervised manner.
The result is the segmentation network $\mathtt{f}_{\theta}$ in \cref{fig:network_diagram}, whose encoder $\mathtt{E}_\theta$ has been initialised from the result of the first step, and then its weights are frozen after this step (indicated by \Snowflake~in \cref{fig:network_diagram}).

\subsubsection{Uncertainty Training}
The objective of this step is to maximise the quality of the model's uncertainty estimates for target domain images.
For this reason, this step uses many unlabelled images from target domains, i.e. domains distinct from the source domain.
This ensures that even if the segmentation quality degrades due to distributional shift, a model is trained to describe the uncertainty quantitatively so that the system as a whole can act appropriately.
The result is the segmentation network $\mathtt{f}_{\phi}$ in \cref{fig:network_diagram}, which shares the same architecture of $\mathtt{f}_{\theta}$, but is parameterised differently.
Prior to training, the encoder $\mathtt{E}_\phi$ has been initialised from the result of the first step, and the decoder $\mathtt{D}_{\phi}$ is randomly initialised.

This final step is our core contribution, and is presented in the next section.


\section{Uncertainty Training with MIM}
During the second step, a network with general weights has been fine-tuned to perform semantic segmentation on the labelled source dataset.
Although performant on the source domain, this network, $\mathtt{f}_\theta$, has lost some of the generality of the pre-trained model; consequently, both the segmentation and uncertainty estimation quality has decreased on \gls{ood} data (see \cref{subsec:results_weight}).

Now, we want to train a new network that can segment the source domain to the same quality as the trained \ftheta but produce higher-quality uncertainty estimates on target domains.
We thus initialise $\mathtt{f}_\phi$ with the encoder from the first step and a random decoder.
This allows us to start with a general model, which can be specialised to segment the source domain well but remain general enough to perform uncertainty estimation on the target domains.
We hypothesise that generality is important for uncertainty estimation because, although it does not require the recognition of all semantic objects, it does require their detection in order to represent them distinctly from the in-distribution classes.

\subsection{Learning Uncertainty Estimation}      \label{subsec:learning_uncertainty}

This work uses a method based on \cite{gammassl}, in which the core of this method are more thoroughly presented.
Without labels, we use the assumption that segmentation consistency over image augmentation approximates ground-truth accuracy -- i.e. if we have two networks that assign two corresponding pixels from two views of the same image to the same class, this is likely to be accurate; otherwise, it is not.

This assumption draws from works in self-supervised learning works \cite{chen2020simple, dinov2,ibot}, where networks are trained under the assumption that \textit{maximising} their consistency over both \gls{candr} and masking augmentations, leads to \textit{maximising} task performance.
Here, the objective is instead to learn to \textit{detect} inconsistency, in order to \textit{detect} when the frozen network exhibits bad task performance.
Unlike \cite{gammassl}, this work defines a masking policy $\mathtt{m}$, which produces masked image $I^m = \mathtt{m}(I)$ (see \cref{subsec:masking_procedure} for details).


Firstly, \fphi segments $I$ and $I^{m}$ to produce $s_{\phi}$ and $s^{m}_{\phi}$ respectively, while \ftheta segments $I$ to produce $s_{\theta}$.
Our aim is to train \fphi such that either: (1) \fphi produces a masked segmentation $s^{m}_{\phi}$ that is consistent with the unmasked frozen model's segmentation $s_{\theta}$, and thus $s_{\theta}$ was likely to have been accurate or (2) $s^{m}_{\phi}$ and $s_{\theta}$ are inconsistent, and therefore one or both segmentations are likely incorrect, \textit{however} \fphi expresses uncertainty with $s_{\phi}$.



Initially, the soft consistency loss is computed between $s^{m}_{\phi}$ and $s_{\theta}$, using the cross-entropy function $\mathcal{H}$, seen as \ding{172} in \cref{fig:network_diagram}.
In order to achieve our stated aim, we allow \fphi to reduce the loss by masking out pixels which it estimates to be uncertain via low values of $\max{[s_{\phi}]}$ (masking depicted as $\boxplus$ in \cref{fig:network_diagram}).

We choose to mask the loss in a hard manner requiring a threshold $\gamma$ to use \cref{eq:mgamma} to calculate $M_\gamma^{\phi}$ (see \ding{173}). 
The calculation of $\gamma$ (see \cite{gammassl} for more detail), is such that the mean value of $M_\gamma^{\phi}$ is equal to the mean value of hard consistency mask $M_c$ (see \ding{174}), where $M_c$ is calculated as: 
\begin{equation}
M_c^{(i)}=
\begin{cases}
1&\text{~if~}\operatorname{argmax}[s_{\theta}^{(i)}] = \operatorname{argmax}[s_{\phi}^{m(i)}]\\
0&\text{otherwise}
\end{cases}
\end{equation}

We calculate $\gamma$ in this way because the method seeks to find pixels where $s_\theta$ is inaccurate, therefore the proportion of certain pixels in $M_\gamma^{\phi}$ should be equal to the proportion of accurate pixels, therefore we use consistency as an approximation.
%
Finally using both \ding{173} and \ding{175}, we define the following objective:
\begin{equation}\label{eq:Lc}
    L_\text{c} =  \frac{\sum_{i}^{NHW} M_\gamma^{\phi (i)}\ \mathcal{H}[\rho_{\text{\tiny{T}}}{(s_{\theta}^{(i)})}, s_{\phi}^{m(i)}]}{\sum_{j}^{NHW}M_\gamma^{\phi (j)}}
\end{equation}

Where $\rho_{\text{\tiny{T}}}(\cdot)$ is a sharpening function \cite{assran2021semi}, where $\text{T}=0.5$.
The latter's inclusion means not only that $L_c$ maximises consistency for certain pixels, but also the increases their certainty, which helps to increase the separation between certain and uncertain pixels.

This uncertainty training is done in conjunction with supervised training on the source domain, such that the total loss is the sum of $L_c$ and the supervised losses found in \cite{mask2former}.

\subsection{Masking Procedure} \label{subsec:masking_procedure}
The masking policy, $\mathtt{m}$, needs to be designed such that segmentations are consistent between \ftheta and $\mathtt{f}_{\phi} \circ \mathtt{m}$ when the mutually assigned class is accurate.
For this reason, masked and unmasked predictions are compared in semantic segmentation space, as opposed to RGB pixel space~\cite{mae}, or an abstract feature space~\cite{dinov2, ibot}.

Defining the masking policy in this work is a significant challenge, due to the nature of the images used.
In any given driving image, there are a large number of different semantic objects and the range in the scale of objects is vast, e.g. the difference in scale between a traffic light in the distance and a car a few metres from the camera.
This is in contrast to datasets typically used in \gls{mim} such as ImageNet \cite{imagenet}, which contain one semantic entity of interest per image, with a smaller variation in scale.

The challenge posed by driving images is two-fold: (1) if an entire small object is masked out, the masked image contains no information from which the model can predict its existence (2) if only small regions are masked out, then the task is too easy, as interpolation can be used to maximise consistency, therefore deteriorating our assumption about consistency approximating accuracy.



Our solution is to use a masking policy that chooses mask elements $M^{(i)}$ independently to produce mask $M \in \{0,1\}^{\hat{H}\times \hat{W}}$, where:
\begin{equation}
M^{(i)} \sim \mathrm{Bernoulli}(p_{\text{mask}})
\end{equation}
where $p_{\text{mask}}=0.5$ unless otherwise stated.
Choosing elements independently reduces the chances of masking out entire semantic objects, and empirically a range of $p_{\text{mask}}$ lead to an appropriate task difficulty (see \cref{subsec:mim_versus_candr}).
Additionally, there were no benefits to more complex masking policies in our experiments, such as learning a masking policy that maximised the inconsistency between unmasked and masked segmentations, or masking only uncertain regions.
Masks are applied as $\mathtt{m}(I) = M \cdot [\mathtt{E}]_{0:1}(I)$, where $[\mathtt{E}]_{0:1}$ is the patch embedding of encoder $\mathtt{E}$, thus the dimensions of the mask are $\hat{H}, \hat{W} = \frac{H}{P}, \frac{W}{P}$, where $P$ is the patch size of $\mathtt{E}$.
Due to the masking scheme's simplicity, our method is both easy to implement and inherently general, as it is not biased to the specific scale or location of the semantic instances in any given dataset.

\section{Experimental Setup} \label{sec:experimental_setup}


\subsection{Network Architecture} \label{subsec:network_arch}
The network architecture for this work follows that of Mask2Former~\cite{mask2former}, comprising an encoder $\mathtt{E}$, a transformer decoder, and a convolutional decoder, grouped in the variable $\mathtt{D}$ in the above discussion.
As in DINOv1 \cite{dinov1} and DINOv2 \cite{dinov2}, the encoder is a DeiT~\cite{deit} transformer.




\subsection{Data}
In this work, we use three types of data.
Firstly, we use labelled images from the source domain.
Secondly, we use unlabelled images from domains distinct from the source domain, called the target domains.
Lastly, we use labelled images from the target domains for testing.

The labelled source domain dataset is Cityscapes \cite{cityscapes} (CS).
As described, the target domains ideally require a large corpus of unlabelled training images and a smaller number of labelled test images.
The SAX Semantic Segmentation benchmark  \cite{gammassl} provides labels for three target domains: London (LDN), New Forest (NF) , and Scotland (SCOT), where this ordering describes an increasing distributional shift w.r.t. Cityscapes.
Additionally, we use the BerkeleyDeepDrive~\cite{bdd} (BDD) dataset as a fourth target domain for training and testing.
WildDash~\cite{wilddash} (WD), a very diverse set of labelled driving images, is used as a test dataset to measure network generalisation, in addition to the validation set of images in Cityscapes (CS).

\subsection{Model Variants} \label{subsec:model_variants}

\paragraph*{Data} Using the \gls{mim} task, we train models on each unlabelled target domain, to evaluate how a model can be trained to be detect error in target, unlabeled domains and how it generalises to unseen target domains.
The target domain used can be found in the model name.

\paragraph*{Initialisation} We initialise models with either DINOv1 or DINOv2 to determine how the feature representation affects uncertainty estimation, under the assumption that DINOv2 is more general than DINOv1 -- as evidenced by its better transfer learning performance to a wider variety of tasks.
Denoted as \texttt{d2} and \texttt{d1} in the model names.

\paragraph*{Input Augmentation} We also compare the models trained with a \gls{mim} task to models trained with \gls{candr} augmentation~\cite{gammassl}, allowing us to investigate the differences between \gls{mim} or \gls{candr} in terms of performance and hyperparameter sensitivity.

\paragraph*{Freezing} (a) For the \gls{candr} task, we investigate the effect of \textit{not} freezing $\mathtt{f}_\theta$, see $\mathtt{f}^\texttt{+}_{\theta}$, (b) For the supervised baselines, $\mathtt{E}$ is frozen.
In contrast to this work, \cite{gammassl} did not use weights from general pretraining and so the representation of the target domain needed to be substantially improved during training. 
Freezing $\mathtt{E}$ is denoted by $\mathtt{E}^*$, and contrasts with networks fully fine-tuned in a supervised manner.

\begin{figure*}[t]
\centering
\includegraphics[width=0.9\textwidth]{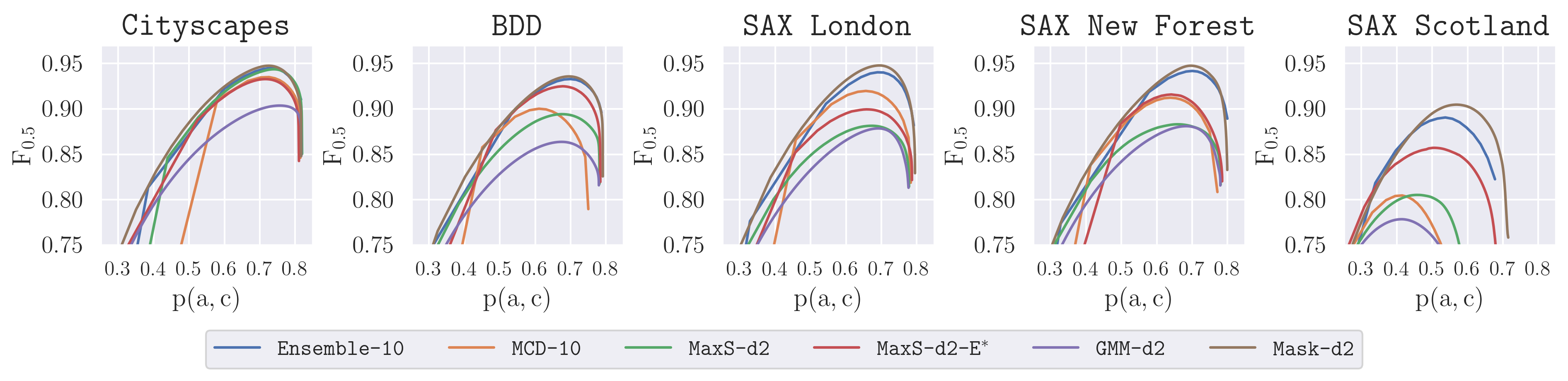}
\caption{
In these plots, we measure misclassification detection performance using $\mathrm{F_{0.5}}$ scores plotted against the proportion of pixels that are certain and accurate $\mathrm{p(a,c)}$.
The baselines are trained only with labelled Cityscapes data, while our proposed model, $\texttt{Mask-d2}$, leverages unlabelled images from the domain in which testing is occurring.
All models are able to perform uncertainty estimation similarly well for Cityscapes, however when tested on the distributionally-shifted target domains, $\texttt{Mask-d2}$'s performance exceeds that of the baselines.
The gap in $\mathrm{MaxF_{0.5}}$ score between $\texttt{Mask-d2}$ and $\texttt{MaxS-d2}$, \texttt{MaxS-d2-E$^{*}$} is descriptive of the benefit of our proposed uncertainty training.
}
\label{fig:fhalf_plot}
\end{figure*}

\begin{table*}[h]
\centering
\resizebox{0.8\textwidth}{!}{
\begin{tabular}{cccccccc}
\hline
& \multicolumn{1}{l}{} & \multicolumn{6}{c}{$\mathrm{MaxF}_{0.5}\text{ @ }\mathrm{p(a,c)}$}        \\
& Method               & CS & LDN & NF & SCOT  & BDD & WD \\ \hline
\multirow{14}{*}{\rotatebox[origin=c]{90}{Baselines}} 

&  \texttt{Ensemble-5-d2}  & 0.944 @ 0.721 & 0.939 @ 0.699 & 0.94 @ 0.697 & 0.894 @ 0.552 & 0.932 @ 0.695 & 0.912 @ 0.624 \\
&  \texttt{Ensemble-10-d2} & 0.945 @ 0.727 & 0.94 @ 0.689 & 0.942 @ 0.701 & 0.891 @ 0.537 & 0.933 @ 0.7 & 0.913 @ 0.626 \\
&  \texttt{MCD-5-d2} & 0.935 @ 0.726 & 0.919 @ 0.656 & 0.912 @ 0.64 & 0.803 @ 0.416 & 0.926 @ 0.679 & 0.903 @ 0.599 \\
&  \texttt{MCD-10-d2} & 0.935 @ 0.726 & 0.92 @ 0.656 & 0.912 @ 0.639 & 0.805 @ 0.415 & 0.926 @ 0.679 & 0.903 @ 0.599 \\
&  \texttt{MaxS-d2-E$^{*}$} & 0.933 @ 0.714 & 0.899 @ 0.658 & 0.916 @ 0.641 & 0.857 @ 0.507 & 0.925 @ 0.678 & 0.908 @ 0.609 \\
&  \texttt{MaxS-d2} & 0.944 @ 0.739 & 0.881 @ 0.674 & 0.883 @ 0.659 & 0.805 @ 0.458 & 0.894 @ 0.674 & 0.87 @ 0.603 \\
&  \texttt{GMM-d2}  & 0.904 @ 0.757 & 0.878 @ 0.692 & 0.881 @ 0.683 & 0.778 @ 0.415 & 0.864 @ 0.674 & 0.805 @ 0.567 \\
&  \texttt{MaxS-d1-E$^*$} & 0.912 @ 0.67 & 0.859 @ 0.579 & 0.899 @ 0.612 & 0.82 @ 0.359 & 0.886 @ 0.586 & 0.86 @ 0.483 \\
&  \texttt{MaxS-d1} & 0.936 @ 0.716 & 0.858 @ 0.61 & 0.887 @ 0.628 & 0.79 @ 0.343 & 0.896 @ 0.622 & 0.853 @ 0.497 \\
&  \texttt{GMM-d1} & 0.894 @ 0.751 & 0.789 @ 0.639 & 0.823 @ 0.61 & 0.687 @ 0.281 & 0.827 @ 0.656 & 0.743 @ 0.572 \\
&  \texttt{C\&R-NF-d2} & 0.928 @ 0.703 & 0.911 @ 0.641 & 0.925 @ 0.662 & 0.869 @ 0.518 & 0.912 @ 0.669 & 0.891 @ 0.61 \\
&  \texttt{C\&R-NF-d2-f}$^\texttt{+}_{\theta}$  & 0.931 @ 0.69 & 0.894 @ 0.658 & 0.902 @ 0.66 & 0.863 @ 0.524 & 0.923 @ 0.676 & 0.908 @ 0.618 \\
&  \texttt{C\&R-NF-d1} & 0.917 @ 0.666 & 0.867 @ 0.567 & 0.901 @ 0.607 & 0.825 @ 0.364 & 0.898 @ 0.618 & 0.874 @ 0.522 \\
&  \texttt{C\&R-NF-d1-f}$^\texttt{+}_{\theta}$ & 0.919 @ 0.67 & 0.894 @ 0.599 & 0.907 @ 0.612 & 0.752 @ 0.338 & 0.891 @ 0.622 & 0.852 @ 0.5 \\
 \hline
 \multirow{6}{*}{\rotatebox[origin=c]{90}{Ours}} & \texttt{Mask-LDN-d2}    & 0.945 @ 0.735 & \textbf{0.948} @ 0.693 & \textbf{0.948} @ 0.697 & 0.889 @ 0.481 & 0.934 @ 0.694 & 0.913 @ 0.605 \\
& \texttt{Mask-NF-d2} & \textbf{0.947} @ 0.724 & 0.941 @ 0.679 & \textbf{0.948} @ 0.696 & 0.903 @ 0.517 & 0.927 @ 0.677 & 0.902 @ 0.582 \\
& \texttt{Mask-SCOT-d2}  & 0.934 @ 0.706 & 0.924 @ 0.645 & 0.939 @ 0.693 & \textbf{0.905} @ 0.568 & 0.92 @ 0.662 & 0.899 @ 0.579 \\
& \texttt{Mask-BDD-d2}   & 0.938 @ 0.725 & 0.931 @ 0.675 & 0.942 @ 0.698 & 0.891 @ 0.501 & \textbf{0.936} @ 0.696 & \textbf{0.918} @ 0.631 \\
& \texttt{Mask-LDN-d1}   & 0.931 @ 0.693 & 0.9 @ 0.623 & 0.919 @ 0.641 & 0.848 @ 0.361 & 0.91 @ 0.63 & 0.879 @ 0.508 \\
& \texttt{Mask-SCOT-d1} & 0.926 @ 0.679 & 0.881 @ 0.599 & 0.919 @ 0.636 & 0.851 @ 0.397 & 0.907 @ 0.624 & 0.87 @ 0.502 \\
\hline
\end{tabular}
}
\caption{
Misclassification Detection performance described by $\mathrm{MaxF}_{0.5}\text{ @ }\mathrm{p(a,c)}$ for a range of test domains.
\label{tab:fhalf}
}
\end{table*}

\subsection{Baselines}

We compare our method to baselines from uncertainty estimation and \gls{ood} detection literature.
Firstly, we consider epistemic-uncertainty estimation methods, Monte Carlo Dropout~\cite{dropout} and \cite{ensembles}, both initialised with DINOv2, and trained on Cityscapes.
These methods are significantly slower to run than both ours and \gls{ood} detection methods, and thus cannot be the solution for many mobile robotics applications, however represent a gold-standard for uncertainty estimation.

The chosen \gls{ood} methods take a model trained in a supervised manner on the source domain and leverage the learned feature representation by designing an inference procedure that produces an OoD score.
The inference procedures chosen involve: (1) a max-softmax score \cite{maxsoftmax} (\texttt{MaxS}), and the Mahalanobis distance between mean features from the source dataset and extracted feature from a target domain~\cite{mahalanobis} (\texttt{GMM}).
In addition to supervised training, these networks are initialised with DINOv1 and DINOv2.

\subsection{Metrics}
The quality of uncertainty estimation is evaluated through a misclassification detection task.
This is a binary classification problem, whereby the positive and negative label states are \textit{accurate} and \textit{inaccurate} respectively, and the positive and negative prediction states are \textit{certain} and \textit{uncertain} respectively.
The ideal model classifies all accurate pixels as certain, and inaccurate pixels as uncertain. 

This allows us then to use metrics typically used for binary classification, namely area under PR curves ($\mathrm{AUPR}$) and maximum $\mathrm{F}_\beta$ scores, which we report with the proportion of pixels that are both accurate and certain, $\mathrm{p(a,c)}$.
Due to the above definitions, a model that prioritises precision over recall puts a higher negative cost on pixels that are (\textit{certain}, \textit{inaccurate}).
In the context of safety-critical robotics, this is key, resulting in our decision to measure the $F_{0.5}$ score, i.e. a score that prioritises precision over recall.


\section{Results} \label{sec:results}
\subsection{MIM for different target domains}
The model with the highest quality uncertainty estimates is our $\texttt{Mask-d2}$ networks trained with unlabelled data from same domain as testing.
This is evidenced in \cref{fig:fhalf_plot}, \cref{tab:fhalf} and \cref{tab:aupr}, in which the former two consider the point at which uncertain pixels are optimally rejected, and the latter considers the full sweep of uncertainty thresholds.

On top of this, the results show that the $\texttt{Mask-d2}$ models also generalise effectively to unseen target domains.
The $\texttt{Mask-BDD-d2}$ model performed best on the diverse unseen WildDash benchmark, with other $\texttt{Mask-d2}$ models similar in performance to the epistemic approaches.
For every target domain, the $\texttt{Mask-d2}$ models outperform the baselines based on a fully fine-tuned segmentation network ($\texttt{MaxS-d2}$,$\texttt{GMM-d2}$), demonstrating how the uncertainty training maintains generality for improved uncertainty estimation.

\begin{table}[h]
\centering
\resizebox{\columnwidth}{!}{
\begin{tabular}{cccccccc}
\hline
& \multicolumn{1}{l}{} & \multicolumn{6}{c}{$\mathrm{AUPR}$}        \\
& Method               & CS & LDN & NF & SCOT  & BDD & WD \\ \hline
\multirow{14}{*}{\rotatebox[origin=c]{90}{Baselines}} 

&  \texttt{Ensemble-5-d2}  & 0.98 & 0.976 & 0.979 & 0.933 & 0.967 & 0.948 \\
&  \texttt{Ensemble-10-d2} & 0.981 & 0.976 & 0.978 & 0.932 & 0.967 & 0.946 \\
&  \texttt{MCD-5-d2} & 0.977 & 0.971 & 0.963 & 0.778 & 0.971 & 0.959 \\
&  \texttt{MCD-10-d2}  & 0.977 & 0.971 & 0.963 & 0.786 & 0.971 & 0.959 \\
&  \texttt{MaxS-d2-E}$^*$ & 0.975 & 0.958 & 0.968 & 0.931 & 0.97 & 0.961 \\
&  \texttt{MaxS-d2}  & 0.982 & 0.933 & 0.942 & 0.861 & 0.94 & 0.919 \\
&  \texttt{GMM-d2} & 0.936 & 0.894 & 0.914 & 0.838 & 0.896 & 0.845 \\
&  \texttt{MaxS-d1-E}$^*$ & 0.966 & 0.92 & 0.953 & 0.889 & 0.942 & 0.925 \\
&  \texttt{MaxS-d1} & 0.98 & 0.913 & 0.942 & 0.859 & 0.949 & 0.918 \\
&  \texttt{GMM-d1} & 0.921 & 0.781 & 0.866 & 0.741 & 0.835 & 0.739 \\
&  \texttt{C\&R-NF-d2} & 0.972 & 0.967 & 0.975 & 0.936 & 0.954 & 0.937 \\
&  \texttt{C\&R-NF-d2-f}$^\texttt{+}_{\theta}$ & 0.978 & 0.927 & 0.932 & 0.92 & 0.965 & 0.96 \\
&  \texttt{C\&R-NF-d1} & 0.972 & 0.929 & 0.954 & 0.883 & 0.951 & 0.935 \\
&  \texttt{C\&R-NF-d1-f}$^\texttt{+}_{\theta}$ & 0.972 & 0.949 & 0.957 & 0.794 & 0.941 & 0.915 \\
 \hline
 \multirow{6}{*}{\rotatebox[origin=c]{90}{Ours}} & \texttt{Mask-LDN-d2}     & 0.985 & \textbf{0.987} & \textbf{0.985} & 0.947 & \textbf{0.973} & 0.966 \\
& \texttt{Mask-NF-d2}  & \textbf{0.988} & 0.984 & \textbf{0.985} & 0.958 & 0.964 & 0.958 \\
& \texttt{Mask-SCOT-d2}   & 0.977 & 0.975 & 0.98 & \textbf{0.96} & 0.965 & 0.957 \\
& \texttt{Mask-BDD-d2}   & 0.98 & 0.977 & 0.981 & 0.952 & 0.972 & \textbf{0.967} \\
& \texttt{Mask-LDN-d1}   & 0.979 & 0.956 & 0.969 & 0.91 & 0.957 & 0.937 \\
& \texttt{Mask-SCOT-d1}   & 0.977 & 0.936 & 0.966 & 0.92 & 0.959 &  0.933 \\
\hline
\end{tabular}
}
\caption{
Misclassification Detection performance summarised over all possible thresholds described by $\mathrm{AUPR}$
\label{tab:aupr} for a range of test domains.
}
\end{table}

\subsection{MIM with different weight initialisations} \label{subsec:results_weight}
Each \texttt{Mask-d2} model outperforms the trained \texttt{Mask-d1} models, even when the \texttt{Mask-d1} model is trained on the same domain as testing.
Additionally, when the encoder was not fine-tuned in \texttt{Mask-E$^*$}, this also relates to better uncertainty estimation.

The higher scores for \texttt{Mask-d2} over \texttt{MaxS-d2-E$^*$}, and \texttt{Mask-d1} over \texttt{MaxS-d1-E$^*$} demonstrate that our uncertainty training has effectively improved the representation beyond that of general pretraining, by including task-specific information, and without the over-specialisation seen in \texttt{MaxS-d2}.

\subsection{MIM versus C\&R} \label{subsec:mim_versus_candr}
In our tests in the NF domain, $\texttt{Mask}$ models are generally superior to the \texttt{C\&R} models, while also exhibiting additional benefits.
We trained two \texttt{Mask} and \texttt{C\&R} variants with different augmentation hyperparameters.
We use $p_{\text{mask}}=[0.25, 0.75]$, one variation of \texttt{C\&R} with less colour-space augmentation, and the other cropping using different scale parameters.
This experiment demonstrated that the $\mathrm{MaxF}_{0.5}\text{ @ }\mathrm{p(a,c)}$ = [$0.926\text{@}0.655$, $0.891\text{@}0.648$] for the \texttt{C\&R} in the order presented, and similarly for \texttt{Mask} models [$0.945\text{@}0.705$, $0.951\text{@}0.713$] on the NF dataset.
The scores for \texttt{C\&R} vary considerably more than for \texttt{Mask}, thereby making masking more convenient. 

\subsection{Freezing $\mathtt{f}_\theta$}
For the $\texttt{C\&R}$ task and training on NF, when DINOv1 is used and $\mathtt{f}_\theta$ is not frozen, we see a boost in uncertainty estimation performance in NF over the equivalent with frozen \ftheta.
This however does not necessarily hold for other domains and DINOv2 -- this shows that $\mathtt{f}_\theta$ can be frozen when using sufficiently general representations, removing the possibility of feature collapse seen in \cite{gammassl}.



\section{Conclusion} \label{sec:concl}
This work presents a method that leverages \gls{mim} with unlabelled images and general feature representations in order to train a network to jointly perform high-quality semantic segmentation and uncertainty estimation.
We show that this method outperforms a number of baselines on a number of different test datasets, alongside investigations into the effect of initial representations, augmentations and training datasets.

\clearpage
\bibliographystyle{IEEEtran}
\bibliography{biblio}

\clearpage
\onecolumn

\renewcommand{\textfraction}{0.05}
\renewcommand{\topfraction}{0.95}
\renewcommand{\bottomfraction}{0.95}
\renewcommand{\floatpagefraction}{0.35}
\setcounter{totalnumber}{5}

\captionsetup[figure]{justification=centering}

\section{\Large{Appendix}}     \label{appendix}
\begin{figure}[h]
    \centering
    \begin{subfigure}{\textwidth}
        \includegraphics[width=\textwidth]{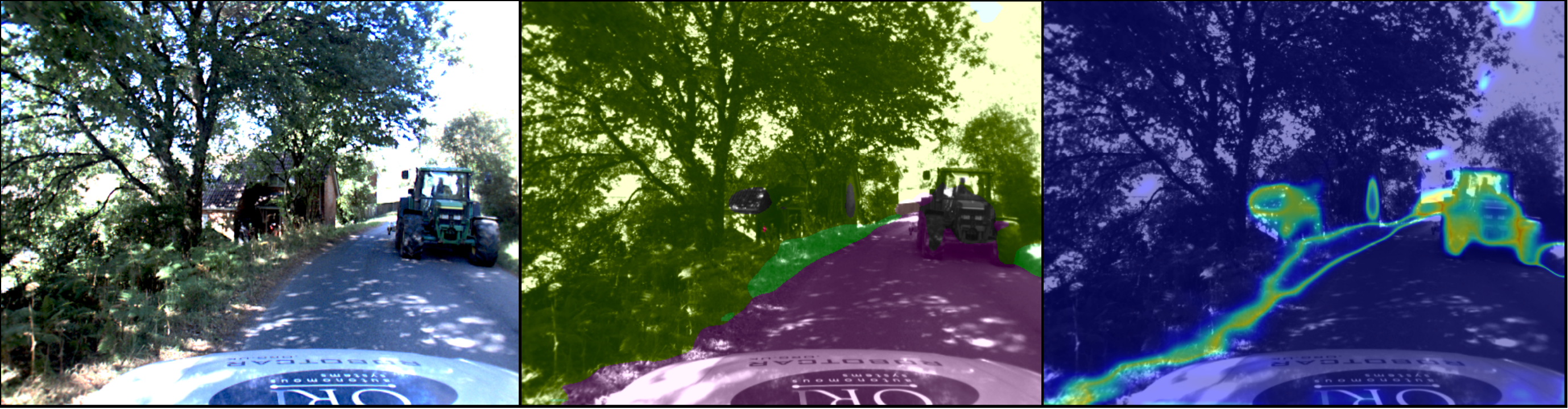}
        \caption{SAX New Forest}
    \end{subfigure}
    
    \begin{subfigure}{\textwidth}
        \includegraphics[width=\textwidth]{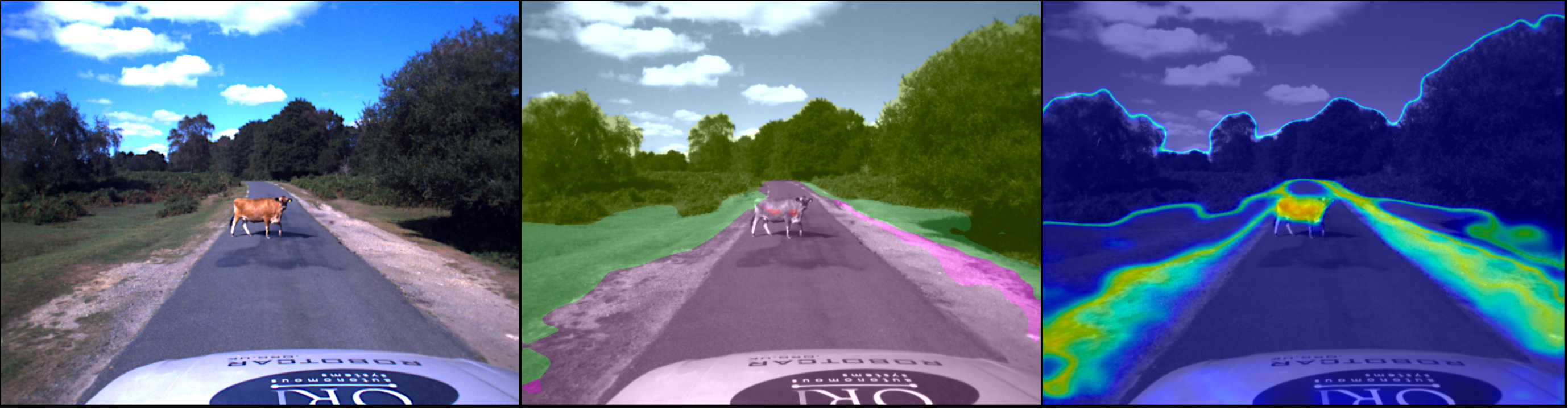}
        \caption{SAX New Forest}
    \end{subfigure}    
    
    \begin{subfigure}{\textwidth}
        \includegraphics[width=\textwidth]{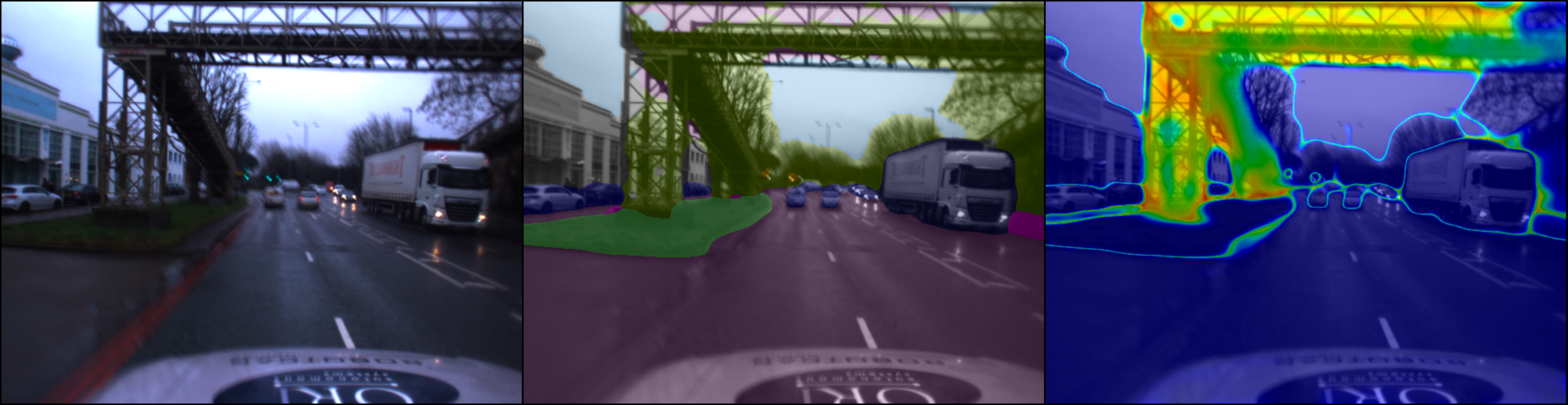}
        \caption{SAX London}
    \end{subfigure}
    
    \begin{subfigure}{\textwidth}
        \includegraphics[width=\textwidth]{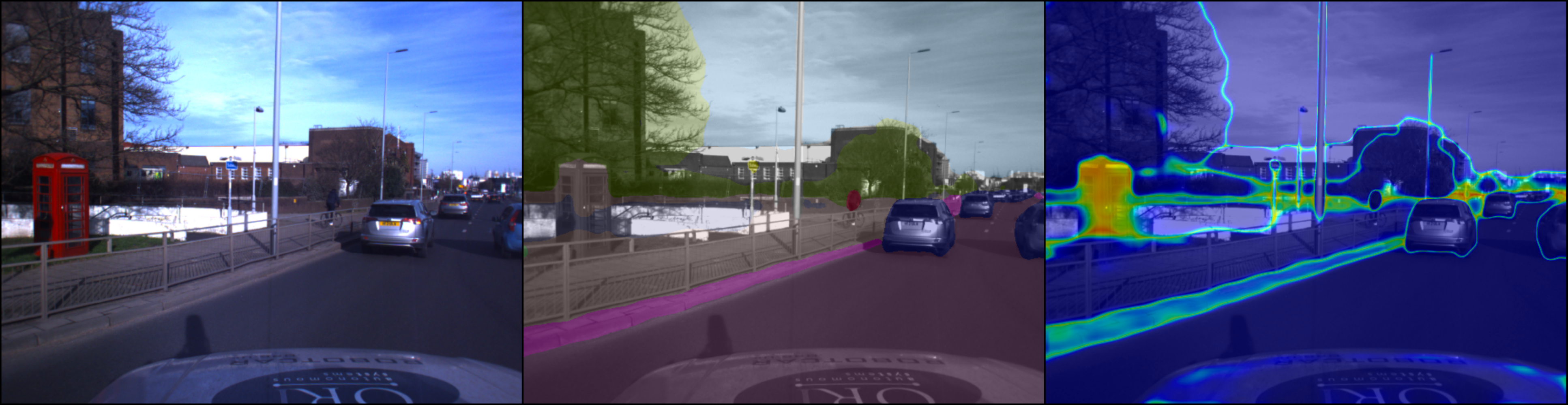}
        \caption{SAX London}
    \end{subfigure}    
    \captionsetup{justification=raggedright,singlelinecheck=false}
    \caption{
    Qualitative results for the proposed \texttt{Mask-d2} model, presenting (left) an RGB image, (middle) the semantic segmentation and (right) the estimated uncertainty in the $\mathtt{jet}$ colour map, where red is uncertain and blue is certain.
    For each distributionally-shifted image, the incorrect segmentations are effectively detected by the model's estimated uncertainty.
    The variant of \texttt{Mask-d2} model used is that which was trained on the same domains as the test image shown (see caption).
    }
    \label{fig:mgssl_qual_results}
\end{figure}

\end{document}